\documentclass[conference]{IEEEtran}
\IEEEoverridecommandlockouts
\usepackage{cite}
\usepackage{amsmath,amssymb,amsfonts}
\usepackage{algorithmic}
\usepackage{graphicx}
\usepackage{textcomp}
\usepackage{xcolor}
\usepackage{float}
\def\BibTeX{{\rm B\kern-.05em{\sc i\kern-.025em b}\kern-.08em
    T\kern-.1667em\lower.7ex\hbox{E}\kern-.125emX}}
\begin{document}

\title{Unveiling Public Perceptions: Machine Learning-Based Sentiment Analysis of COVID-19 Vaccines in India
%{\footnotesize \textsuperscript{*}Note: Sub-titles are not captured in Xplore and should not be used} \thanks{Identify applicable funding agency here. If none, delete this.}
}

\author{\IEEEauthorblockN{1\textsuperscript{st} Milind Gupta}
\IEEEauthorblockA{\textit{Dept. of Computing Science \& Mathematics} \\
\textit{Dundalk Institute of Technology}\\
Dundalk, Ireland \\
milindgupta@outlook.com}
\and
\IEEEauthorblockN{2\textsuperscript{nd} Abhishek Kaushik}
\IEEEauthorblockA{\textit{Dept. of Computing Science \& Mathematics} \\
\textit{Dundalk Institute of Technology}\\
Dundalk, Ireland \\
abhishek.kaushik@dkit.ie}
%\and
% \and
% \and
% \and
 }

\maketitle

\begin{abstract}
In March 2020, the World Health Organisation declared COVID-19 a global pandemic as it spread to nearly every country. By mid-2021, India had introduced three vaccines: Covishield, Covaxin, and Sputnik. To ensure successful vaccination in a densely populated country like India, understanding public sentiment was crucial. Social media, particularly Reddit with over 430 million users, played a vital role in disseminating information. This study employs data mining techniques to analyze Reddit data and gauge Indian sentiments towards COVID-19 vaccines. Using Python's Text Blob library, comments are annotated to assess general sentiments. Results show that most Reddit users in India expressed neutrality about vaccination, posing a challenge for the Indian government's efforts to vaccinate a significant portion of the population.
\end{abstract}


\section{Introduction}
% \par The World Health Organization (WHO), on 30 January 2020 designated an outbreak of never seen before novel corona virus in human population. Later, on 11 March 2020 this disease, termed as COVID-19 was declared a pandemic by WHO. As of April 2023, there were total 762,201,169 confirmed cases and 6,889,743 confirmed deaths globally \cite{who}, out of which there were 44,179,712 current active cases and 530,916 confirmed deaths in India \cite{mohfw}. 
% There are about 13,321,840,096 vaccines administered globally \cite{who} out of which 220,66,16,373 vaccine doses have been given in India alone \cite{mohfw}. As per Indian Council of Medical Research, until the end of July 2022, about 95\% of the adult population was vaccinated with first dose whereas about 80\% population was vaccinated with two doses \cite{icmr}. 
\par According to WHO, vaccines hesitancy is one of the major hurdle that the world faces today \cite{doi:10.1080/21645515.2020.1780846}. In order to make vaccination successful in a populous and diverse country like India, sentiments of people should be taken into account for. To gather the public opinions towards vaccines, classical surveys are often used \cite{PMID:33082575} but they are costly and tedious. Whereas, social media analysis can be cheaper, quicker and more practical to collect large set of real-time data \cite{info:doi/10.2196/32335}. Nowadays, in order to find the general opinion of the public, there is no better place other than social media. 
% Founded in March 2005 by Steve Huffman and Alexis Ohanian, Reddit is one such social media platform that has seen remarkable growth since its inception. The name \emph{Reddit} is a play-on-words with the phrase \emph{read it} \cite{Reddit}. 
Reddit is one of the most popular social media platform with an estimation of 56 million daily active users \cite{RedditUsers} who are referred to as \textit{Redditors}. Reddit is organized into thousands of individual communities called \textit{subreddits}. Each subreddit focuses on a specific topic or theme, such as technology, news, movies, gaming, science, art, or fitness. Users can subscribe to subreddits that interest them and see posts related to those topics on their front page. For this research, subreddits like \emph{r/IndiaSpeaks}, \emph{r/indianews}, \emph{r/COVID19} and \emph{r/india} are considered. User comments related to keywords like covishield, covaxin, sputnik, vaccination etc. are extracted from these subreddits and sentiment analysis is performed over them.
\par Sentiment Analysis is a study to analyze and evaluate the sentiments, opinion, thoughts and user's emotions which they express on social media platform \cite{bdcc3030037}. Sentiment Analysis can also be defined as a method to understand the view of people on various topics and eventually finding their emotional orientation towards those topics \cite{bdcc4010003}. There are generally three major emotional dimensions that are used to evaluate the sentiments, namely \emph{positive, negative and neutral} \cite{RAO2016978}.

% Sentiment Analysis is an approach to Natural Language Processing which involves the use of data mining, machine learning and artificial intelligence techniques to mine the opinions and categorize them into positive, negative or neutral \cite{TechTarget}.
% As of May 2023, over 950 million people in India are completely vaccinated, whereas the percentage of people fully vaccinated in USA, France and Russia are 69\%, 78\% and 55\% respectively \cite{OurWorldinData}. The key to a successful COVID-19 vaccination program is getting people to accept and take the vaccine. There are many reasons why people might not want to get vaccinated, such as concerns about the safety or effectiveness of the vaccine, or simply not believing that they need it. ECDC (European Centre for Disease Prevention and Control) is working to identify and address these barriers by providing accurate information about vaccines, working with communities to understand their concerns, and supporting research into the factors that influence vaccine acceptance \cite{ECDC}.
  
India is a diverse country and most of the people speak more than one language. Young population of India is more multilingual than their elder generation. About 52\% of Indian youth is bilingual while 44\% of Indian urban population speaks two languages \cite{TimesofIndia}. With such diversity in languages, the opinions expressed by Indian people on social media are of mixed-languages with Hinglish (Hindi written in Roman script) being the most prominent one \cite{shah2019sentiment}. An example for Hinglish sentence is \emph{Is vaccine ki efficacy kya hai?} which translates to \emph{What is the efficacy of this vaccine?}, here \emph{vaccine} and \emph{efficacy} are English words whereas \emph{is, ki, kya \& hai} are Hindi words (Refer Table \ref{tab:table123}). There are number of challenges faced during Sentiment Analysis like emotion detection, context detection and sarcasm detection. Working on multi-lingual data like Hinglish is even greater challenge because of scarcity of resources available \cite{bdcc3030037}.
\begin{table}[!ht]
    \centering
    \begin{tabular}{|p{0.8in}|p{0.8in}|p{0.5in}|p{0.5in}|}
    \hline
        Comment & Translation & English Words & Hinglish Words \\ \hline
        Is vaccine ki efficacy kya hai? & What is the efficacy of this vaccine? & vaccine, efficacy & is, ki, kya, hai \\ \hline
    \end{tabular}

\caption {Hinglish Comment Example}
\label{tab:table123}
\end{table}
% \begin{figure}[H]
%     \centering
%     \includegraphics[width=\columnwidth]{hinglish_comment.jpg}
%     \caption{Hinglish Comment Example}
%     \label{Fig.9}
% \end{figure}

COVID-19 vaccines have been developed using different technologies (Table \ref{tab:table1} shows the list of vaccines approved for emergency use.). Here are some of the main technologies used for COVID-19 vaccines:
 \begin{enumerate}
     \item \textbf{mRNA:} mRNA vaccines instruct your body to produce a protein that stimulates an immune response, without employing a live virus. For e.g. Pfizer/BioNTech and Moderna \cite{HSE.ie}.
     \item \textbf{Viral vector:} Viral vector vaccines act as messengers, employing a modified form of another virus to transport instructions to your body's cells. For e.g. AstraZeneca and Janssen \cite{HSE.ie}.
     \item \textbf{Protein subunit:} The vaccine consists of a variant of a protein present on the outer surface of the COVID-19-causing virus. Additionally, it incorporates an adjuvant (ingredient) that aids in enhancing the immune responses to the vaccine. For e.g. Novavax \cite{HSE.ie}.
     \item \textbf{Inactivated:} This particular vaccine variant includes an inactivated SARS-CoV-2 virus, which is identified by the immune system to trigger a response without inducing COVID-19 infection. This response contributes to the development of immune memory, empowering your body to combat future encounters with SARS-CoV-2. For e.g. Sinovac and Covaxin \cite{BSI}.
 \end{enumerate}

The rest of the paper is organized as follows. Subsection \ref{moti} describes the motivation behind the study on sentiment analysis of vaccines in India. Section \ref{relate} presents the related works in the field of sentiment analysis. This section is further divided into 4 different sections like Worldwide sentiment analysis of COVID-19 vaccines, Sentiment Analysis of COVID-19 vaccines in India, Sentiment Analysis of COVID-19 Vaccines in other countries with multi-language data and Sentiment Analysis of mixed language posts on Social Media in India. Section \ref{meth} is about the methodology which provides the understanding of data exploration and preliminary analysis performed. Section \ref{result} summarizes the results obtained from number of machine learning models run on the dataset. %Section \ref{future} provides information on future approaches of the study. 
Future work and conclusion of the study can be found in Section \ref{conc}.

%%%%%%%%%%%%%%%%%%%%%%%%%%%%%%%%%%%%%%
%%%%%%%%%%%%%%%%%%%%%%%%%%%%%%%%%%%%%%
\subsection{Motivation}
\label{moti}
COVID-19 cases in India peaked in May 2021 with more than 300,000 daily cases but by the end of the year 2021, the total daily cases were reduced to 6000 \cite{IndiaCovid}. The first vaccine was administered in India on 16$^{th}$ January 2021 with the launch of World's largest vaccination roll-out \cite{UNICEF}. In a country of 1.3 billion people, such large scale vaccination drive was required to bring down the daily COVID-19 cases by 98\%. As stated by Richard Baldwin, a professor of international economics at the Graduate Institute in Geneva: \emph{"This virus is as economically contagious as it is medically contagious"} \cite{ForeignPolicy} hence it is mandatory to study the sentiments of the general public towards vaccination and also to make AI trained system in medical facilities to fight against the pandemics quickly. 
Mermer, G. et al. \cite{mermer} in their research have stated that the reluctance and resistance towards vaccines have an adverse impact on the delivery of immunization services. The promotion of anti-vaccination campaign and the normalization of anti-vaccination are eroding the trust of the public in healthcare and posing a threat to public health. 
During the Polio Vaccination program in India, the government faced a number of challenges to get children vaccinated. It took more than 30 years of efforts to make India a Polio free country in 2014. \par Bellatin A. et al. \cite{Bellatine005125} in their research \emph{"Overcoming vaccine deployment challenges among the hardest to reach: lessons from polio elimination in India"} explained one of the primary obstacles to eradicate polio during its last stages was the resistance from communities hesitant to take the vaccine. In Uttar Pradesh, such reluctance to immunization stemmed from a history of mistrust between the government and its vulnerable populations \cite{Coates68}. These populations in India are defined as marginalized subgroups requiring additional support, care, and customized interventions due to historical, structural, and institutional barriers. This pertains to communities like migrant laborers, religious minorities, and socioeconomically disadvantaged sectors of society. These groups were hesitant about the vaccination drive, primarily because they perceived other problems such as the lack of basic amenities like sanitation and food security to be more pressing concerns. Due to a history of mistrust, some members of religious minorities who were impacted by government family planning programmes in the 1970s were unwilling to vaccinate their children in the future \cite{PDI}. Therefore, such sentiments analyses are required to understand the sentiments of such vulnerable groups of people and change their mindset towards vaccines in order to achieve better public health.
\par
 There have been numerous studies carrying out sentimental analysis on COVID-19 vaccinations. Among them very few have taken Indian perspective and even lesser studies have taken multi-language texts into consideration. This study's objectives are to (1) investigate the sentiments of the people of India with regard to the COVID-19 vaccine, and (2) investigate the machine learning process and approach to automating the sentiment analysis process. If successful, the findings of this study will assist the health industry and the government in better understanding the sentiments of the people and in better planning their vaccination campaigns.

During the exploratory study, we formed following research questions mentioned in next subsection.
\begin{table*}[!ht]
  \centering
   \begin{tabular}{|p{1.5in}|p{1.5in}|p{1.5in}|p{1.5in}|}
   \hline
       Common Name & Technology & Origin Country & First Authorization \\ \hline \hline
       Oxford–AstraZeneca & Adenovirus vector & United Kingdom, Sweden & December 2020 \\ \hline
       Pfizer–BioNTech & RNA & Germany, United States & December 2020 \\ \hline
       Janssen (Johnson \& Johnson) & Adenovirus vector & United States, Netherlands & February 2021 \\ \hline
       Moderna & RNA & United States & December 2020 \\ \hline
         Sinopharm BIBP & Inactivated & China & July 2020 \\ \hline
        Sputnik V & Adenovirus vector & Russia & August 2020 \\ \hline
        CoronaVac & Inactivated & China & August 2020 \\ \hline
        Novavax & Subunit/virus-like particle & United States & December 2021 \\ \hline
        Covaxin & Inactivated & India & January 2021 \\ \hline
        Valneva & Inactivated & France, Austria & April 2022 \\ \hline
        Sanofi-GSK & Subunit & France, United Kingdom & November 2022 \\ \hline
        Sputnik Light & Adenovirus vector & Russia & May 2021 \\ \hline
   \end{tabular}
\caption {COVID-19 vaccines authorized for emergency use for full use \cite{vaccines}}
\label{tab:table1}
\end{table*}
\subsubsection{Research Questions}
\begin{itemize}
    \item \textbf{RQ1.} Is it feasible to apply machine learning techniques to automate sentiment analysis on health-related data?
    \item \textbf{RQ2.} What is the overall sentiment of the COVID-19 vaccination among Indian people?
\end{itemize}

For this study Google Scholar was used to perform the exploratory analysis. Research papers related to COVID-19 vaccinations were selected. The research papers that were selected were based on the following study topics:
\begin{itemize}
    \item Worldwide Sentiment Analysis of COVID-19 Vaccines.
    \item Sentiment Analysis of COVID-19 Vaccines in India.
    \item Sentiment Analysis of COVID-19 Vaccines in other countries with multi-language data.
    \item Sentiment Analysis of mixed language posts on Social Media in India.  
\end{itemize}

\section{Related Works}
\label{relate}
The related work has been segregated based on different levels like Worldwide, India specific and countries other than India. \textit{Google Scholar} is a valuable tool for researchers and academics seeking to explore research articles related. By searching for keywords like \textit{"sentiment analysis"}, \textit{"sentiment analysis covid 19 vaccines"} and \textit{"sentiment analysis covid 19 vaccines in india"} we were able to access a comprehensive database of scholarly articles, research papers, and publications from various sources. We searched for articles published after 2020, and Google Scholar displayed over 35,000 results. These research papers were then categorized into four different categories as mentioned in section \ref{moti}.
\subsection{Worldwide Sentiment Analysis of COVID-19 Vaccines}
 \par Lazarus, J.V. et al. \cite{PMID:33082575} surveyed the potential acceptance of COVID-19 vaccine globally. A survey was conducted where more than 13000 participants from 19 countries responded to a 22 items questionnaire. The items contained questions related to trust in pandemic information, vaccine uptake and other standard demographic questions regarding age, gender, income etc \cite{lazarus2020covid}. 71.5\% responded positively to take a vaccine it were proven safe and effective. Countries with more than 80\% vaccine acceptance were Asian nations like China, South Korea and Singapore. There was also comparatively high acceptance in middle-income countries Brazil, India and South Africa.
\par Marcec R et al. \cite{marcec2022using} conducted sentiment analysis of AstraZeneca, BioNTech and Moderna vaccine using data retrieved from Twitter API. The analysis focused only on English language tweets from 1 December 2020 to 31 March 2021. %A tool named \emph{AFINN lexicon} \cite{conf/msm/Nielsen11} which is specifically designed to perform sentiment analysis on micro-blogging sites such as Twitter, was used. The tool contains around 2477 words with a value (-5 (highly negative) to +5 (highly positive)) assigned to them. A total of 701891 tweets were retrieved and included in the daily sentiment analysis. The retrieved data was tokenised using \emph{tidytext} package of R programming language and merged with AFINN lexicon to calculate the average daily sentiment. To compare the changes in sentiment for each vaccine over time statistical analysis was conducted in R-programming. 
\par Ansari MTJ et al. \cite{ansari} have used Naïve Bayes Theorem (using TextBlob library of python) for the categorization. For data retrieval from Twitter API, a python package Tweepy was used. After cleaning the dataset by removing non-English words, stop words and emoticons, around 820,000 tweets were categorized into \emph{positive, negative} and \emph{neutral} using Bayes theorem. 
\par Chen et al. \cite{10.1145/3240508.3240533} have focused on Twitter Sentiment Analysis via emojis embedding. In this study, they have proposed two mechanism, namely Word-guide Attention-based LSTM (Long Short Term Memory) and Multi-level Attention-based LSTM. The study involved following steps: initialization of bi-sense emoji embedding (assigning two distinct tokens to each emoji, one is for the particular emoji used in positive contexts and the other one is for this emoji used in negative contexts), generating senti-emoji embedding, and sentiment classification via the proposed attention-based LSTM networks.
\par Puri N et al. \cite{doi:10.1080/21645515.2020.1780846} in their study have researched on impact of social media on vaccine hesitancy. The study discusses on growing public concerns on the impact of anti-vaccination content available across social media platforms. substantial amount of anti-vaccine content is shared on social media. Betsch et al. \cite{doi:10.1177/1359105309353647} and Nan et al. \cite{doi:10.1080/10410236.2012.661348} have demonstrated in their study that anti-vaccine content shared on social media platforms or blogs has negative impact on people to vaccinate. It is thus important for influential people in politics or medical science to correctly disseminate fact-based information on social media to encourage people towards vaccination \cite{Zhange025866}.
\par Bonifazi et al. \cite{BONIFAZI2022103095} have presented a new approach to analyzing discussions on social networks. Their approach is based on multilayer networks, which allow them to study the interactions between users at different levels. The authors compared their approach to single-network approaches and found that the multilayer network approach is more effective at capturing the complexity of social interactions.  the authors analyzed the pro-vax, neutral, and anti-vax discussions on Twitter and found that the anti-vax community is more tightly knit than the other two communities.
%\par Francesco et al. \cite{info:doi/10.2196/42227} have amassed nearly 300 million tweets in the English language pertaining to COVID-19 vaccines over a span of 12 months, employing a comprehensive list of over 80 pertinent keywords. The objective of the study was to identify misinformation using two approaches. The initial method detected tweets that contained hyperlinks to websites with diminished credibility. These websites had been annotated by journalists, fact-checkers, and media specialists for frequently disseminating misinformation, unfounded rumors, conspiracy theories, baseless assertions, highly biased propaganda, clickbait, and similar content. To identify YouTube videos shared on Twitter that might contain misinformation, the authors extracted unique video identifiers from the links shared in the tweets. They then used the YouTube API's Videos:list endpoint to query the status of each video.
%%%%%%%%%%%%%%%%%%%%%%%%%%%%%%%%%%%%%%%%%%%%%%%%%%%%%%%%%%%%%%%%%%%%
\subsection{Sentiment Analysis of COVID-19 Vaccines in India}
Praveen et al. \cite{praveen2021analyzing} analyzed the attitude of Indian citizens towards COVID-19 vaccines. The data was collected from Twitter using python library \emph{Twint}. More than 73000 English language tweets were selected for the analysis. Their study comprises of two different methods, namely Sentiment Analysis using Natural Language Processing and Latent Dirichlet Allocation (LDA). Former method uses TextBlob library to analyze each word of the documents presented in corpus and categorize them into positive, negative or neutral categories. Latter method, LDA uses \emph{Bag of Words} assumption. The premise on which it operates is that certain groups of words tend to be linked with particular subjects.
\par To indicate the negative or positive sentiments in the tweets retrieved from Twitter API, Ponmani, K et al. \cite{Ponmani_Thangaraj_2022} have used NLP and improved random forest classification algorithm. Tweets containing the keywords like \emph{Covishield}, \emph{Covaxin}, \emph{Sputnik} and \emph{COVID-19} were collected. Data pre-processing involved steps like Stop Words removal, Tokenization and Normalization (using Stemming and Lemmatization). TF-IDF Vectorization process was used to translate tokens into numbers because algorithms works on numbers not texts. Clustering using MEEM \emph{(Modified Efficient Expectation Maximization)} technique along with improved Random Forest (the count of trees in the random forests grows iteratively) were used to categorize the data for each vaccine into positive, negative and neutral.
% \par Mudassir et al. \cite{9544512} have used three NLP approaches namely, ABSA \emph{(Aspect Based Sentiment Analysis)}, VADER \emph{(Valence Aware Dictionary for Sentiment Reasoning)} and TextBlob to understand the vaccine perception in India. The tweets in this paper have been classified in the categories of negative, neutral and positive. For Hinglish tweets, the labelling has been done manually. It was analysed that ABSA models performs better in terms of accuracy.
% \par Another method to perform sentiment analysis is demonstrated by Chinnasamy P. et al. \cite{CHINNASAMY2022448} using NLTK \emph{Natural Language Toolkit} package. In this study, the NLTK library was used for text pre-processing, tokenization, NLTK Sentiment Intensity Analyzer and lexicon based Emotional Analysis. After data pre-processing, sentiment analysis was done using decision tree algorithm. The Decision Tree algorithm was chosen because it is faster and more accurate than the Naïve Bayes algorithm \cite{prasad2020framework}.
%%%%%%%%%%%%%%%%%%%%%%%%%%%%%%%%%%%%%%%%%%%%%%%%%%%%%%%%%%%%%%%%%%%%
\subsection{Sentiment Analysis of COVID-19 Vaccines in other countries with multi-language data}
Niu Q. et al. \cite{info:doi/10.2196/32335} analyzed the public opinion of before and at the commencement of COVID-19 vaccination in Japan. COVID-19 Twitter data collected by Georgia State University's Panacea Lab \cite{epidemiologia2030024} was used in the study. Tweets in English and Japanese-English mix language were tokenized and python libraries SpaCy and GiNZA were used to remove English and Japanese stop words. AWS tools supporting multiple languages were used to label the corpora into four categories: positive, negative, neutral and mixed. Further Latent Dirichlet Allocation topic modelling was applied on vaccine related tweets. 
\par A study conducted by Villavicencio, C. et al. \cite{info12050204} which reports the sentiment of Filipinos towards COVID-19 vaccines. RapidMiner search operator was used in the study to search tweets in Philippines with English and Tagalog languages. After data cleaning (including data annotation) and pre-processing, data was categorize using Naïve Bayes theorem. Model's performance was evaluated using \emph{k-fold cross validation} model. The result showed 81.77\% accuracy.
% \par Nezhad et al. \cite{nezhad2022twitter} conducted study on sentiment analysis of Iranian citizen on home grown and foreign vaccines. More than 800,000 Persian language tweets posted from 1 April 2021 to 30 September 2021 related to the keywords were retrieved from Twitter API. The pre-trained model based on CNN-LSTM architecture was used to assign polarity scores \emph{(positive, negative and neutral)} to each tweet in Persian language \cite{Bokaee_Nezhad_Deihimi_2019}. In this model CNN is used as feature extractor for LSTM. NLP technique \emph{Word2vec} is used for word embedding. The study suggested slightly more negative sentiments towards imported vaccines whereas negative and positive sentiments came out be same for home-grown vaccine.
% \par To find the sentiments of Indonesian people towards COVID-19 vaccines Ritonga, Mulkan, et al. \cite{ritonga2021sentiment} used Naïve Bayes algorithm. Data extraction was done using RapidMiner technique. In data pre-processing cleaning, stop words removing, tokenization and stemming was done. Later feature extraction was done using vectorization (TF-IDF). And eventually model was categorized and evaluated using  Naïve Bayes Classifier. 
%%%%%%%%%%%%%%%%%%%%%%%%%%%%%%%%%%%%%%%%%%%%%%%%%%%%%%%%%%%%%%%%%%%%

\subsection{Sentiment Analysis of mixed language posts on Social Media in India}
Kazhuparambil et al. \cite{kazhuparambil2020classification} in their study \emph{Classification of Malayalam-English Mix-Code Comments using Current State of Art}, used three phase approach for classification of YouTube comments on certain cooking channels. First of the three phase is text classification using the BERT language model. Second phase is the application of machine learning models and final phase involves building a Multi-Layer Perceptron. Malyalam-English mix-code comments were extracted from Youtube API and each record was manually labeled based on the sentiment it shows. The labels were inspired from Kaur et al. \cite{bdcc3030037} and Shah et al. \cite{shah2020opinion}. Multiple vectorizing techniques such as Count Vectorizer, TF-IDF Vectorizer, Hashing Vectorizer etc. were used for Feature Engineering. Among the many models used in the study, XLM was top performing model with 67.32\% accuracy.
\par Shah et al. \cite{shah2020opinion} analyzed the opinions on Marglish (Marathi language + English language mix) and Devanagri script comments on YouTube cooking channels. About 42,551 comments from certain Marathi Cookery Channels were scraped using YouTube API. Next the data was manually labelled into 5 categories. After vectorization and tokenization of the data, parametric and non-parametric models like \emph{Logistic Regression, Decision Tree, Naïve Bayes, Random Forest, Support Vector Machine and Multilayer Perceptron} were used. To evaluate the results, Cross-Validation, Performance Tuning and statistical Testing was done. Multilayer Perceptron with Count vectorizer provided the best accuracy of 62.68\% on the Marglish dataset.
% \par Pravalika A. et al. \cite{8204074} have used two approaches for the sentiment analysis of Facebook posts related to the movies. First approach is Lexicon-based approach where dictionaries of words annotated with their semantic orientation (polarity) are used to classify the text. Second approach is machine learning approach which involves data collection, data preprocessing, feature creation and lastly, application of machine learning techniques to train the classifier. The machine learning model resulted in the accuracy of 72\% whereas lexicon-based model worked with the accuracy of 86\%.

\section{Methodology}
\label{meth}
The methodology for this research can be divided into four main steps: data collection, data cleaning, annotation, and sentiment analysis. 
% Data collection is the process of gathering the text data that will be used for the analysis, Data cleaning is the process of removing any errors or inconsistencies in the data, Annotation is the process of assigning sentiment labels to each piece of text in the dataset and Sentiment analysis is the process of using machine learning algorithms to classify the sentiment of each piece of text in the dataset.
\begin{table}[!ht]
    \centering
    \begin{tabular}{|l|l|}
    \hline
        Polarity Score & Sentiment \\ \hline \hline
        0 & Neutral \\ \hline
        0 to 0.3 & Weakly Positive \\ \hline
        0.3 to 0.6 & Positive \\ \hline
        0.6 to 1 & Strongly Positive \\ \hline
        -0.3 to 0 & Weakly Negative \\ \hline
        -0.6 to -0.3 & Negative \\ \hline
        -1 to -0.6 & Strongly Negative \\ \hline
    \end{tabular}
\caption {Polarity vs Sentiments}
\label{tab:table2}
\end{table}
\subsection{Data Collection}
To collect the data from Reddit API, a Reddit developer account needs to be created. The account will give necessary credentials to access the API. These credentials are used to authenticate the requests to the Reddit API. A python package \textbf{PRAW} was used to retrieve the data. PRAW is an acronym for \textit{Python Reddit API Wrapper} that allows simple access to Reddit's API. \\
Four subreddits were selected to fetch the comments containing the specific keywords. There were around 32000 reddit comments extracted to do the preliminary analysis. Out of 32000 comments, 8571 comments are related to keyword \emph{covishield}, 7380 comments are related to keyword \emph{covaxin}, 4749 comments related to keyword \emph{sputnik} and 11278 comments are related to keyword \emph{vaccination}. Dataset contains attributes like \emph{post id}, \emph{subreddit}, \emph{post title}, \emph{selftext}, \emph{comments}, \emph{score} and \emph{comment time}.

%%%%%%%%%%%%%%%%%%%%%%%
\subsection{Data Cleaning}
Data preprocessing is the process of cleaning and preparing data for analysis. In sentiment analysis, data preprocessing is essential to ensure that the data is accurate and relevant. First step of data preprocessing is \emph{Tokenization}. %Tokenization involves dividing a phrase, sentence, paragraph, or even an entire text document into smaller units, such as separate words or terms. These individual units are known as tokens. Python libraries like NLTK (Natural Language Toolkit) and spaCy are used for tokenization. In the next step, all the words are converted to lowercase. 
Removing the stop words is another important step before performing sentiment analysis. Stop words are commonly used words that are excluded from searches to help index and parse web pages faster. After the removal of stop words from the corpus, normalization of words is done. Normalization involves converting the words to their root form. 
% This can be done by \emph{stemming or lemmatization}. Lemmatization is a language processing technique that simplifies words to their base or root form, known as a lemma \cite{lemmatization}. It helps in reducing different forms of a word to a common form for analysis or comparison. For example, the words \emph{vaccinating, vaccinates, and vaccinated} would all be lemmatized to the base form \emph{vaccinate}. 
Words in our corpus were normalizaed using \emph{WordNetLemmatizer()} function of NLTK library. In the proceeding step, data vectorization was performed. Vectorization in sentiment analysis is the process of transforming text data into numerical arrays that can be used for machine learning models \cite{vectorization}. Two methods Count Vectorization (or Bag-of-Words) and TF-IDF are used in our study. 
% Bag-of-Words defines a dictionary of unique words contained in the text, and then finds the count of each word within each document. It does not preserve the order of the words, and it does not capture any actual meaning of the words. However, it is a simple and fast way to see a distribution of terms and their frequencies. TF-IDF stands for term frequency-inverse document frequency, and it assigns a weight to each word based on how frequently it appears in a document and how rare it is across all documents. It reduces the importance of common words that do not carry much information, and increases the importance of rare words that are more relevant to the sentiment. It also does not preserve the order of the words, but it captures some semantic information.
%%%%%%%%%%%%%%%%%%%%%%%
\subsection{Annotations}
Annotations in sentiment analysis are labels or tags that are assigned to text data to indicate the polarity and subjectivity of the expressed opinions or emotions. The process of human annotation on approximately 32,000 units of text is both time-consuming and resource-intensive. The primary aim of our study was to assess the viability of employing machine learning models to analyze data and efficiently comprehend sentiments in order to gain a deeper understanding and explore our research questions.  Although human annotations are a valuable mechanism that can assist in mitigating model bias in machine learning. Given the time constraints, we made the decision to employ a weak annotation mechanism and utilize a transfer learning strategy in order to annotate our data. This was achieved through the utilization of the text blob Python module. Annotated datasets serve as training data for the models, allowing them to learn patterns and associations between textual features and sentiment. These annotations provide a basis for the models to generalize and make predictions on new, unseen text samples. %Annotations can be \textit{manual} or \textit{automatic}. In manual annotation, human annotators read the text and label it with the appropriate sentiment (positive, negative, neutral). In automatic annotation, the machine learning model learns to label the text without any human intervention.

In this study we have labelled the data into six categories, strongly positive, positive, weakly positive, strongly negative, negative, weakly negative and neutral (Refer Section \ref{senti} for further detail). Table \ref{tab:table2} shows the sentiments assigned to the comments based on the polarity score.
\newline Table \ref{tab:table3} shows the example of comments and their associated sentiments.
\begin{table}[!ht]
    \centering
    \begin{tabular}{|p{2in}|p{0.5in}|}
    \hline
        Comment & Polarity \\ \hline \hline
        Vaccine stopped in 27 countries. But allowed in india. & Negative \\ \hline
        This is excellent. It gives a boost to India's coronavirus vaccination campaign. & Positive \\ \hline
        Are the sputnik vaccines being used for vaccinations in any part of the country?  & Neutral \\ \hline

    \end{tabular}
\caption {Comments \& their sentiments}
\label{tab:table3}
\end{table}
%%%%%%%%%%%%%%%%%%%%%%%
\subsection{Sentiment Analysis}
\label{senti}
A simple sentiment analysis was performed on the extracted data using \textit{TextBlob} library. Polarities of the comments were calculated using \emph{textblob.sentiment.polarity} function. The comments were categorized into 7 different categories namely, \emph{Strongly Positive, Positive, Weakly Positive, Neutral, Weakly Negative, Negative, Strongly Negative}. After categorizing the data based on polarity, we found out that there were about 34\% neutral comments, 32\% weakly positive comments and 16\% weakly negative comments.
Table \ref{tab:table4} shows the frequency of each sentiment in the corpus.
\begin{table}[!ht]
\centering
\begin{tabular}{l l l}
\hline
\textbf{Sentiments} & \textbf{Frequency} & \textbf{Percentage}\\
\hline
Neutral & 10370 & 34.04\% \\
Weakly Positive & 9884 & 32.45\% \\
Positive & 2929 & 9.61\% \\
Strongly Positive & 747 & 2.45\% \\
Weakly Negative & 4894 & 16.07\% \\
Negative & 1343 & 4.41\% \\
Strongly Negative & 296 & 0.97\% \\
\hline
\end{tabular}
\caption{Sentiments Frequency}
\label{tab:table4}
\end{table}

\begin{figure*}[t]
    \centering
    \includegraphics[width=0.8\textwidth]{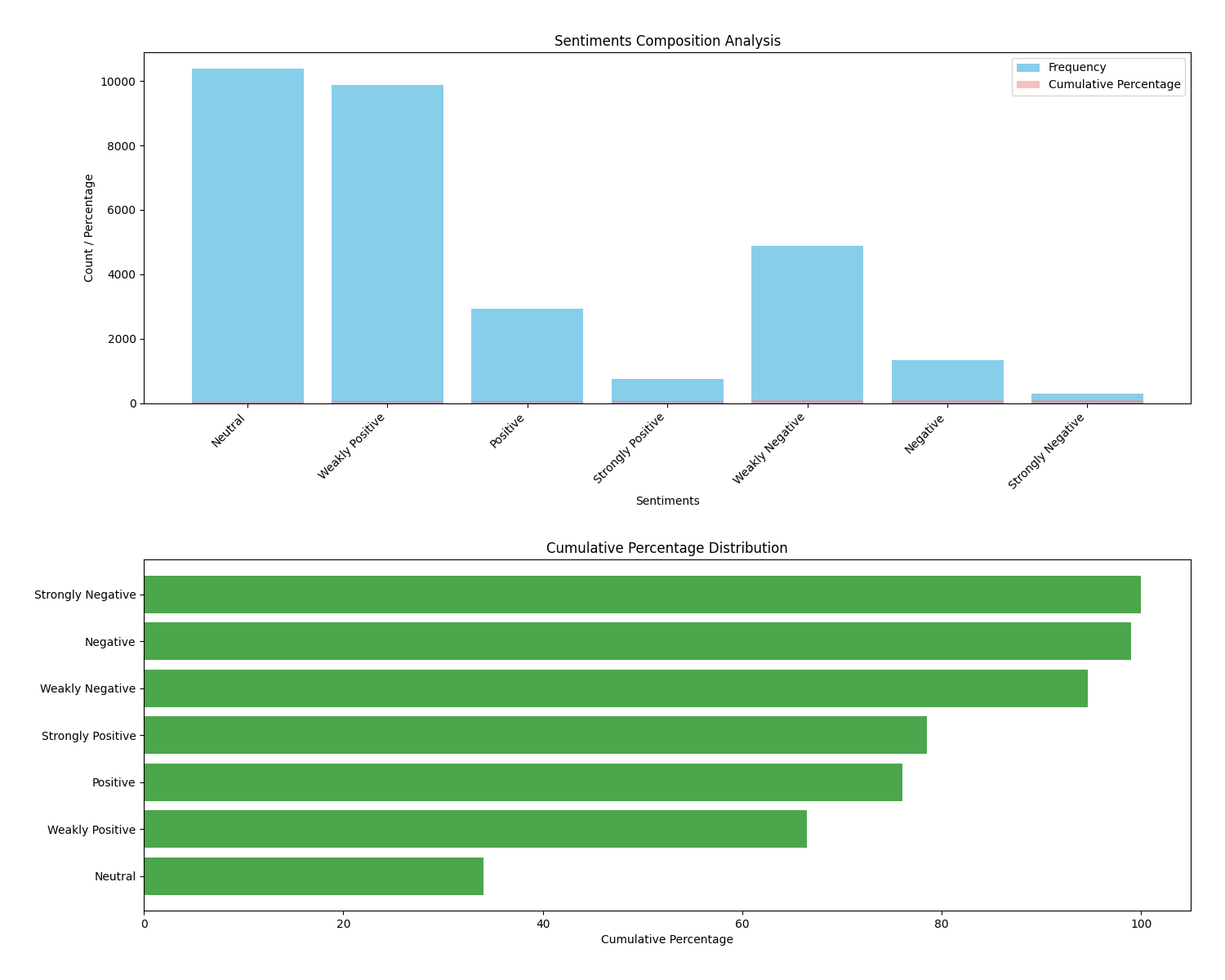}
    \caption{Combined Sentiment Analysis}
    \label{fig:graph}
\end{figure*}

%%%%%%%%%%%%%%%%%%%%%%%%
\subsection{Data Splitting}
Data splitting is the process of dividing a dataset into two or more subsets. This is done in machine learning to evaluate the performance of a machine learning model. The most common way to split data is into a training set and a test set. We have used \emph{train\_test\_split} function of \emph{sklearn} package. Our dataset has been divided into two subsets, where the two subsets train and tests are divided into a ratio of \textbf{7:3} respectively.
\begin{table*}[!ht]

\centering
\begin{tabular}{l l l l l l l}
\hline
\textbf{Vectorizer} & \textbf{Model} & \textbf{Parameter} & \textbf{Accuracy} & \textbf{F1 Score} & \textbf{Precision} & \textbf{Recall}\\
\hline
TF-IDF & KNN & 5 neighbors & 0.38 & 0.26 & 0.66 & 0.38 \\
TF-IDF & KNN & 10 neighbors & 0.37 & 0.24 & 0.75 & 0.37 \\
TF-IDF & KNN & 15 neighbors & 0.37 & 0.23 & 0.75 & 0.37 \\
TF-IDF & KNN & 20 neighbors & 0.36 & 0.22 & 0.75 & 0.36 \\
TF-IDF & Support Vector Machine & Linear & 0.79 & 0.78 & 0.79 & 0.79 \\
TF-IDF & Random Forest & 20 estimators & 0.76 & 0.74 & 0.77 & 0.76 \\
TF-IDF & Random Forest & 40 estimators & 0.77 & 0.75 & 0.78 & 0.77 \\
TF-IDF & Random Forest & 60 estimators & 0.77 & 0.75 & 0.78 & 0.77 \\
Count Vectorizer & KNN & 5 neighbors & 0.46 & 0.38 & 0.6 & 0.46 \\
Count Vectorizer & KNN & 10 neighbors & 0.43 & 0.33 & 0.58 & 0.43 \\
Count Vectorizer & KNN & 15 neighbors & 0.42 & 0.31 & 0.58 & 0.42 \\
Count Vectorizer & KNN & 20 neighbors & 0.41 & 0.29 & 0.57 & 0.41 \\
Count Vectorizer & Support Vector Machine & Linear & 0.81 & 0.81 & 0.81 & 0.81 \\
Count Vectorizer & Random Forest & 20 estimators & 0.76 & 0.75 & 0.77 & 0.76 \\
Count Vectorizer & Random Forest & 40 estimators & 0.77 & 0.76 & 0.78 & 0.77 \\
Count Vectorizer & Random Forest & 60 estimators & 0.78 & 0.76 & 0.78 & 0.77 \\
\hline
\end{tabular}
\caption{Non-Parametric Models}
\label{tab:table6}
\end{table*}
\subsection{Machine Learning}
After splitting the data, a number of machine learning models were run on the dataset and there results were recorded.

\subsubsection{K-Nearest Neighbor}
%The K-Nearest Neighbors or KNN algorithm is a non-parametric, supervised machine learning algorithm that can be used for both classification and regression tasks. %It is a simple algorithm that works by finding the k most similar instances (neighbors) to a new instance and then predicting the label of the new instance based on the labels of its neighbors. 
% The textbf{k} value in the KNN algorithm is a hyperparameter that controls how many neighbors are used to make a prediction. The higher the value of k, the more neighbors are used and the smoother the decision boundary becomes. However, a higher value of k can also lead to overfitting.
% \textbf{Euclidean Distance Formula:}
% \begin{equation}
%     \label{eq:euclid}
%     \boxed{d(\mathbf{x},\mathbf{y}) = \sqrt{\sum_{i=1}^n (x_i - y_i)^2}}
% \end{equation}

To perform the KNN classification, an optimum value of parameter \textbf{k} has to be selected. We ran the model on 4 different values of k (5,10,15,20) and best result was selected. The model gave highest accuracy for k = 5.

\subsubsection{Naive Bayes Algorithm}
The Naive Bayes algorithm is a supervised machine learning algorithm that is used for classification tasks. It is based on Bayes' theorem, which is a formula for calculating the probability of an event occurring given the probability of other events occurring \cite{NaiveBayes}
% The Naive Bayes algorithm assumes that the features of a data point are independent of each other. This means that the presence of a particular feature does not affect the probability of another feature being present. This assumption is often not true in real life, but it makes the Naive Bayes algorithm very simple to implement and fast to train. \\
% \textbf{Bayes Theorem:}
% \begin{equation}
%     \label{eq:bayes}
%     \boxed{P(A|B) = \frac{P(B|A)P(A)}{P(B)}}
% \end{equation}
We have used different types of Naive Bayes algorithms in our research, namely, \textbf{Gaussian}, \textbf{Multinomial}, \textbf{Complement} and \textbf{Bernoulli}.

Out of these algorithms, Bernoulli Naive Bayes Algorithm gave best results with 58\% accuracy for both Count and TF-IDF vectorizers.

\subsubsection{Support Vector Machine}
A support vector machine (SVM) is a supervised machine learning algorithm that can be used for both classification and regression tasks \cite{SVM}. %It is a discriminative classifier, which means that it tries to find a decision boundary that separates the two classes of data points.
% The SVM algorithm works by finding the hyperplane that best separates the two classes of data points. The hyperplane is a line or a plane in the feature space that has the maximum margin between the two classes. The margin is the distance between the hyperplane and the closest data points of each class \cite{SVM}.
% There are two types of SVMs:
% \begin{itemize}
%     \item \textbf{Simple SVM:} A simple SVM is a support vector machine (SVM) that uses a linear kernel. This means that the hyperplane that the SVM finds is a straight line in the feature space.
%     \item \textbf{Kernel SVM:} A kernel SVM is a support vector machine (SVM) that uses a kernel function to map the data points into a higher-dimensional feature space. This allows the SVM to find a hyperplane in the higher-dimensional space that can separate the two classes of data points, even if the data is not linearly separable in the original space.
% \end{itemize}

In our research we have used Linear SVM, which gave accuracy of 79\% and 81\% for TF-IDF and Count vectorizer respectively.

\subsubsection{Random Forest Classifier}
Random forest is a machine learning algorithm that is known for its versatility and accuracy. It works by combining multiple decision trees  \cite{RFC}. %Each decision tree makes its own prediction, and the final prediction is made by taking the majority vote of the trees. This helps to reduce the chances of overfitting and improve the accuracy of the model.
% \par Decision trees are a type of supervised machine learning algorithm that can be used for both classification and regression tasks. They work by recursively partitioning the data into smaller and smaller subsets until each subset is homogeneous. The decision tree is then used to make predictions by traversing the tree from the root node to a leaf node, based on the values of the features.
% \par Random forest is a popular supervised machine learning algorithm that can be used for both classification and regression tasks. It is a type of ensemble learning algorithm, which means that it combines multiple decision trees to make predictions. The more trees in the forest, the more robust and accurate the model will be \cite{RFC}

The accuracy for 40 and 60 estimators came out to be same for both TF-IDF and Count vectorizers as shown in Table \ref{tab:table6}. %$The accuarcy when TF-IDF vectorizer was used was 77\%, whereas when Count vectorizer was used, the accuracy was 78\%. 

\subsubsection{Stochastic Gradient Descent}
Gradient descent is a powerful optimization algorithm that can be used to find the minimum or maximum of a function. The gradient descent algorithm works by iteratively moving in the direction of the steepest descent, until it reaches a minimum \cite{SGD}.
%\par Typically, there are three types of Gradient Descent, \textbf{Batch Gradient Descent}, \textbf{Stochastic Gradient Descent} and \textbf{Mini-batch Gradient Descent}.
% In our study we have used Stochastic Gradient Descent. Stochastic gradient descent is a type of gradient descent that uses only a single data point or a small subset of data points to update the model parameters at each iteration. This makes SGD computationally efficient, especially for large datasets. SGD works by iteratively updating the model parameters in the direction of the negative gradient of the cost function.
% % \par The main difference between SGD and batch gradient descent is that SGD only uses a single data point or a small subset of data points to update the model parameters at each iteration. This makes SGD computationally efficient, especially for large datasets. However, it can also be less accurate than batch gradient descent, which uses all of the data points to update the model parameters at each iteration. \\

%%%%%%%%%%%%%%%%%%%%%%%%%%%%%%%%

\begin{table*}[!ht]

\centering

\begin{tabular}{l l l l l l l}
\textbf{Vectorizer} & \textbf{Model} & \textbf{Parameter} & \textbf{Accuracy} & \textbf{F1 Score} & \textbf{Precision} & \textbf{Recall}\\
\hline
TF-IDF & Naive Bayes & Gaussian & 0.44 & 0.46 & 0.55 & 0.44 \\
TF-IDF & Naive Bayes & Multinomial & 0.52 & 0.44 & 0.59 & 0.52 \\
TF-IDF & Naive Bayes & Complement & 0.57 & 0.56 & 0.61 & 0.57 \\
TF-IDF & Naive Bayes & Bernoulli & 0.58 & 0.53 & 0.58 & 0.58 \\
TF-IDF & Stochastic Gradient Descent & N/A & 0.66 & 0.61 & 0.69 & 0.66 \\
Count Vectorizer & Naive Bayes & Gaussian & 0.42 & 0.44 & 0.54 & 0.42 \\
Count Vectorizer & Naive Bayes & Multinomial & 0.58 & 0.56 & 0.63 & 0.58 \\
Count Vectorizer & Naive Bayes & Complement & 0.54 & 0.54 & 0.57 & 0.54 \\
Count Vectorizer & Naive Bayes & Bernoulli & 0.58 & 0.53 & 0.57 & 0.58 \\
Count Vectorizer & Stochastic Gradient Descent & N/A & 0.78 & 0.77 & 0.78 & 0.78 \\
\hline
\end{tabular}
\caption{Parametric Models}
\label{tab:table7}
\end{table*}
%\subsubsection{Cross Validation}

\begin{table}[!ht]
\centering
\begin{tabular}{l l l}
\hline
\textbf{Models} & \textbf{Vectorizer} & \textbf{Score}\\
\hline
Support Vector Machine & TF-IDF & 0.838 \\
Random Forest & TF-IDF & 0.825 \\
Support Vector Machine & Count & 0.818 \\
Stochastic Gradient Descent & Count & 0.849 \\
\hline
\end{tabular}
\caption{Cross Validation Scores}
\label{tab:table5}
\end{table}

\section{Results And Discussion}
\label{result}
We ran a number of machine learning models (as discussed in Section \ref{meth}) on our data and compared the outcomes from each of them. We used two different vectorization techniques and ran multiple machine learning model on each of them. Refer tables \ref{tab:table6} and \ref{tab:table7} for detailed results of each model. We have also explore K- fold cross-validation mechanisms on the best wto models. 
K-fold cross-validation is a statistical method used to evaluate the performance of a machine learning model. It works by dividing the dataset into k folds, or subsets. The model is then trained on k-1 folds and evaluated on the remaining fold. This process is repeated k times, with each fold being used as the test set once. The final evaluation score is the average of the k scores. 
% K-fold cross-validation is a more robust way to evaluate a model than simply splitting the data into a training set and a test set. This is because it helps to avoid overfitting, which is a problem that occurs when the model learns the training data too well and is unable to generalize to new data \cite{KCV}. \\
% Steps involved in k-fold cross-validation:
% \begin{enumerate}
%     \item Divide the dataset into k folds.
%     \item For each fold:
%     \begin{itemize}
%         \item Train the model on the k-1 folds.
%         \item Evaluate the model on the remaining fold.
%     \end{itemize}
%     \item Calculate the average of the k scores.
% \end{enumerate}

\par The table \ref{tab:table5} below shows the cross validation score of Top-2 models each in parametric and non-parametric categories. %Cross Validation was done on these models by taking number of folds as 4 and best result has been added in the table below.

Referring to Table \ref{tab:table4}, we can infer that opinion of Indian public on Reddit was mostly neutral. Our study of reddit comments from Indian citizens found that 34.04\% of the tweets had neutral attitudes, while 2.45\% were strongly positive and 0.97\% were strongly negative. This suggests that there is a significant portion of the population that is undecided about whether or not to get vaccinated. The government and public health officials in India are aware of the challenges posed by the neutral sentiment towards COVID-19 vaccines. They are working to address these challenges by providing more information about the vaccines, increasing access to vaccination, and building trust with the public. It is important to remember that the goal is not to get everyone to agree with vaccination, but rather to get as many people vaccinated as possible. By addressing the concerns of the public and providing them with the information they need, the government and public health officials can help to increase vaccination rates and protect the health of the population. As depicted in the Figure \ref{fig:graph}, there exist two main visual depictions. The upper subplot displays a stacked bar chart that clearly depicts the distribution of sentiments, along with their respective cumulative percentages. It is noteworthy to mention that a significant proportion of the attitudes stated exhibit neutrality and mild positivity, constituting approximately 66\% of the overall distribution of moods. The lower subplot presents a horizontal bar chart that visually represents the cumulative percentage distribution of sentiments. The visualisation provided in this study serves to validate the prevalence of neutral and somewhat positive attitudes, providing a concise overview of the distribution of sentiments as a whole. These visualisations offer valuable insights into the structure of sentiment. The stacked bar chart provides a comprehensive representation of the sentiment distribution. The collection primarily consists of neutral and mildly positive sentiments, which collectively represent around two-thirds of the entire dataset.The resulting cumulative percentage curve illustrates the concentration of attitudes, stressing the prevalence of neutral and somewhat favourable thoughts. The graphical representation of the data underscores the significance of neutral and mildly positive emotions, as these categories have the highest cumulative percentages. This observation further emphasises the substantial impact they have on the general dissemination of emotions.

The sentiment analysis performed helped us in answering the Research Questions we formed in section \ref{moti}. \\

\begin{itemize}
    \item \textbf{RQ1.} Is it feasible to apply machine learning techniques to automate sentiment analysis on health-related data?
        \newline Indeed, "yes" would serve as the response to this particular set of research question. The initial weakly annotated data, obtained using a transfer learning approach, has been used to assess the effectiveness of the approach. The results indicate an accuracy of over 80\% on cross-validated models.  Based on the conducted inquiry, we have examined both parametric and non-parametric algorithms on the health-related data. The data in the present investigation is encoded using a mixed coding language. This problem presents a significant challenge; nonetheless, our chosen strategy appears to hold some promise. In future research, it is recommended to employ the approach of manual annotation to investigate the association between manual automation and annotation based on transfer learning.  Based on our research, it is evident that the utilisation of the transfer learning weak notation technique holds potential for facilitating efficient reference and analysis. This approach has the capacity to support the health industry and government in formulating strategies aimed at enhancing societal well-being.

    \item \textbf{RQ2.} What is the overall sentiment of the COVID-19 vaccination among Indian people?
    \newline
    Based on our investigation to explore this study issue, we have arrived at the conclusion that Indian sentiments are predominantly non-negative, indicating a high likelihood of adaptability.  The prediction of sentiment frequencies obtained in our analysis allows us to confidently assert that a significant proportion of Reddit users adopted a neutral posture in relation to the topic of vaccination. Nevertheless, while taking into account all positive attitudes, whether weak or strong, it can be observed that the Indian population exhibited a greater inclination towards a favourable attitude regarding vaccination rather than a negative one.  The prevailing mood among the Indian population regarding the various available vaccines appears to largely exhibit a neutral stance. The observed phenomenon can be ascribed to a combination of variables, including divergent opinions regarding the effectiveness of vaccines, differing degrees of confidence in the healthcare system, and the impact of mass transmission of information across multiple media platforms.  

\end{itemize}

\par
\begin{figure}[!h]
    \centering
    \includegraphics[width=\columnwidth]{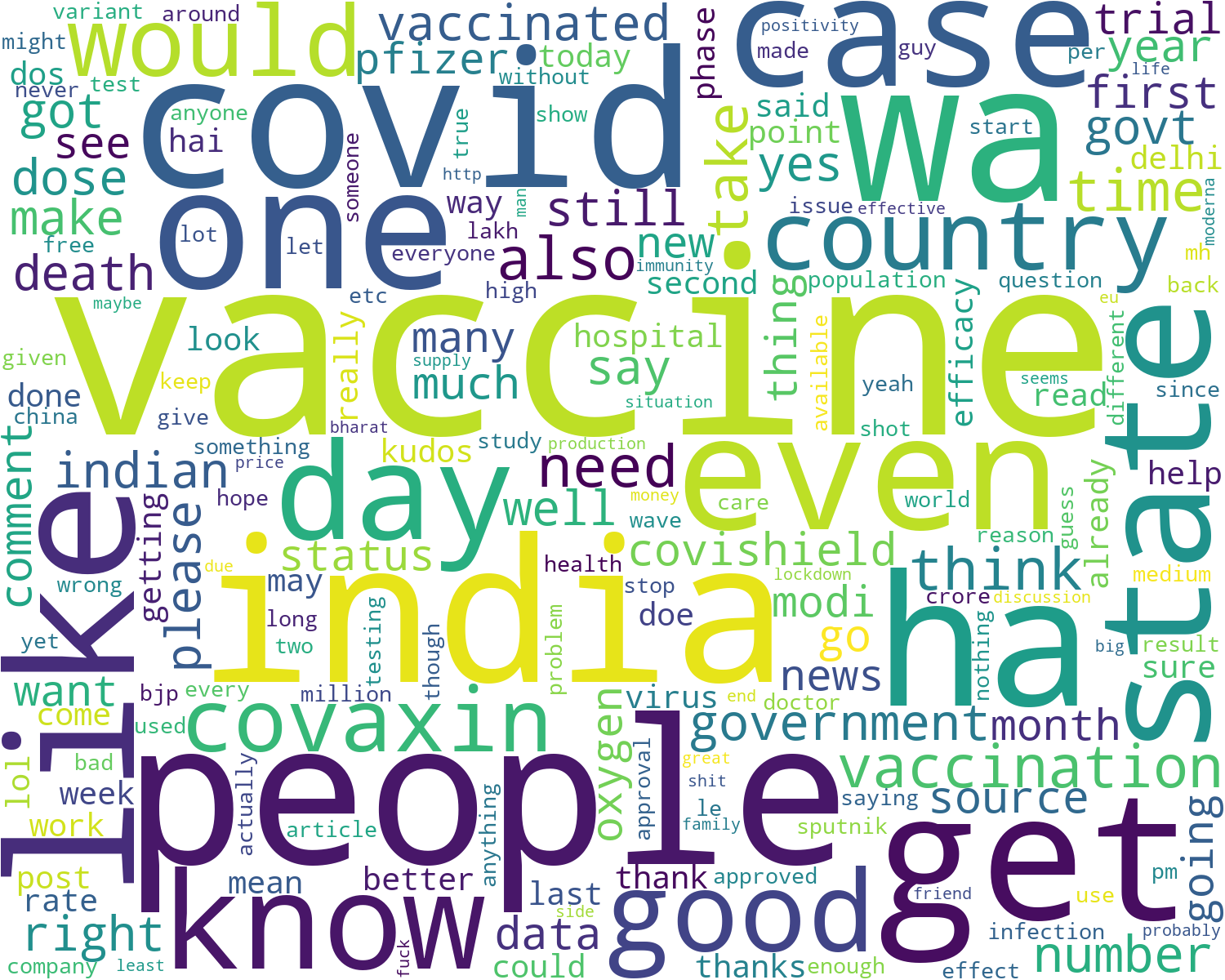}
    \caption{Word Cloud}
    \label{Fig.1}
\end{figure}

 Figure \ref{Fig.1} shows the word cloud. It depicts the cluster of words in the corpus. The bigger and bolder the word appears, the more often it's used in the comments.

% While some individuals might hold positive views about certain vaccines due to their reported effectiveness and safety, others might express skepticism or concerns about potential side effects. On the other hand, there could be people who remain relatively indifferent, possibly due to a lack of substantial information or a sense of uncertainty surrounding the different vaccine options.  \\
%\section{Future Analysis}
%\label{future}

\section{Conclusion \& Future Work}
\label{conc}
This study investigated the attitudes of Indians towards COVID-19 vaccines by classifying them as positive, neutral, or negative. Based on the results obtained, it can be concluded that majority of Indians were neutral towards the vaccines. Only 2.45\% had strongly positive opinion, however, less than 1\% were strongly negative about the vaccines. 
% Around 32000 comments were collected from certain communities on Reddit and this data was pre-processed using techniques like tokenization, removal of stop words, lemmatization and vectorization. The cleaned data was then annotated using textblob library of python. After assigning the polarity to each comment, a number of machine learning models were run on the dataset. Support Vector Machine model and Stochastic Gradient Descent (after count vectorization) gave the best accuracy of 81\%.
%Cross Validation score of Top-2 models for each vectorization method was calculated. 
Cross Validation scores for SVM and Random Forest models after using TF-IDF vectorization were 0.838 and 0.825 and cross validation scores for SVM and Stochastic Graident Descent models after using count vectorization were 0.818 and 0.849.  With the increasing amount of data being generated, the accuracy and reliability of sentiment analysis models is likely to improve. New techniques, such as deep learning, are being developed that can better understand the nuances of human language. In order to get the improved and more accurate results, different research approaches (\textit{Rule-base Approach, Machine Learning Approach, Aspect-based Approach \& Transfer Learning Approach}) can be applied in future.  The limitation of this is the automated annotation. Sentiment analysis models trained without manual annotations may not be able to accurately capture the nuances of human language. For future research, we will be manually annotating the dataset used in this study and comparing the results between manual and automatic annotations. Moreover, it is possible to utilise deep learning or large language models for the purpose of analysing the comments and subsequently training the categorization model. Additionally, we can engage in the examination of Artificial Intelligence explainability models and dedicate efforts towards the assessment of bias and fairness metrics.

%%%%%%%%%%%%%%%%%%%%%%%%%%%%%%%%%%%%%%%%

\bibliographystyle{IEEEtran}
\bibliography{ref}

\end{document}